\documentclass[lettersize,journal]{IEEEtran}

\usepackage{amsmath,amsfonts}
\usepackage{graphicx}
\usepackage{booktabs,multirow}
\usepackage{tabularx}
\usepackage{cite}
\usepackage{url}
\usepackage[colorlinks=true,linkcolor=blue,citecolor=blue,urlcolor=blue]{hyperref}

\usepackage{algorithm}
\usepackage{algorithmic}

\graphicspath{{figs/}{./}}

\begin{document}

\title{Steering the Flow: Inverting Face Recognition Models via Gradient-Guided Flow Matching}

\author{Ye Lu, Shen Wang, Zhaoyang Zhang, Yihan Yan, Li Liu, Runze Liu, and Fanghui Sun
    \thanks{This work was supported in part by Science Challenge Project under Grant TZ2025002, the National Natural Science Foundation of China under Grant 62672145 and Grant 62602158, and in part by the Natural Science Foundation of Heilongjiang Province under Grant LH2024F023.}
    \thanks{Ye Lu, Shen Wang, Zhaoyang Zhang, Yihan Yan, Li Liu, Runze Liu, and Fanghui Sun are with the School of Cyberspace Science, Harbin Institute of Technology, Harbin 150001, China (Corresponding author: Yihan Yan; e-mail: yan.office366.m@hit.edu.cn).}
    \thanks{This manuscript has been accepted for publication in IEEE Internet of Things Journal. This is the author-accepted version of the manuscript. DOI: \href{https://doi.org/10.1109/JIOT.2026.3742815}{10.1109/JIOT.2026.3742815}.}
    \thanks{\copyright\ 2026 IEEE. Personal use of this material is permitted. Permission from IEEE must be obtained for all other uses, in any current or future media, including reprinting/republishing this material for advertising or promotional purposes, creating new collective works, for resale or redistribution to servers or lists, or reuse of any copyrighted component of this work in other works.}
}

\maketitle
\begin{abstract}
Model Inversion Attacks (MIAs) aim to reconstruct representative training samples of target identities from face recognition models, exposing critical security vulnerabilities. Existing methods typically rely on indirect guidance or highly stochastic guidance, making it difficult to stably optimize generation trajectories toward target facial images. In this paper, we propose Steering Flow Model Inversion (SFMI), a novel two-stage white-box model inversion method that reformulates inversion as a trajectory-steering task.
Specifically, Step I, Learning a Generic Flow Matching Prior, pre-trains a generic unconditional Flow Matching model to encode the manifold of human faces as a robust prior. Step II, Attacking with Progressive Guidance Scheduler (PGS), injects time-dependent target-specific gradients during sampling. By backpropagating through the target model to obtain gradients from intermediate generated states, PGS progressively injects adaptive guidance signals into the vector field. This process effectively steers the current generative flow from random noise toward the high-density regions of the target class.
Under an identity-disjoint cross-evaluation setting using the CelebA dataset, SFMI achieves an ACC of 0.9248, an FID of 22.61, and an LPIPS of 0.3874 on the ArcFace target.
Extensive experiments on multiple target models demonstrate that SFMI achieves competitive state-of-the-art performance in attack success and visual fidelity under the evaluated white-box protocol.
\end{abstract}

\begin{IEEEkeywords}
Model inversion attack, face recognition, privacy attack, deep learning.
\end{IEEEkeywords}

\section{Introduction} \label{sec:intro}

Deep Neural Networks (DNNs) have emerged as the backbone of modern biometric authentication, achieving superhuman accuracy in tasks such as Face Recognition (FR)~\cite{wang_survey_2022}. The efficacy of these models is predicated on their capacity to extract high-level semantic features from massive-scale datasets. However, this reliance on extensive training data engenders a fundamental tension between utility and privacy. Theoretical and empirical evidence suggests that DNNs tend to unintentionally ``memorize'' specific training samples rather than merely learning generalized patterns. This phenomenon, coupled with the pervasive deployment of FR systems in critical infrastructure, poses severe privacy risks. In light of stringent regulations such as the GDPR, auditing the potential leakage of sensitive biometric information has become a paramount research imperative.

\begin{figure}[t]
  \centering
  \includegraphics[width=\linewidth]{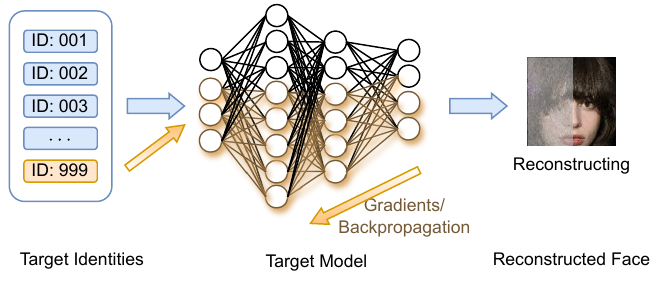}
  \caption{Basic concept of the Model Inversion Attack (MIA).}
  \label{fig:mia_concept}
\end{figure}

Among the emerging privacy threats, Model Inversion Attacks (MIAs) reveal that DNNs---especially face recognition models---are inherently vulnerable to identity leakage. In an MIA, the adversary reconstructs a representative facial image of a target identity by exploiting the information encoded in a trained model. Generally, MIAs are categorized into black-box and white-box settings based on the adversary's knowledge. While the black-box setting restricts access to model queries, the white-box setting assumes full access to the model's parameters and architecture. In this paper, we focus on the white-box scenario, as illustrated in Fig.~\ref{fig:mia_concept}. Here, the adversary leverages full access to the model parameters to backpropagate gradients for a specific label (e.g., ``ID: 999''). This feedback iteratively guides random noise to evolve into a facial image that semantically aligns with the target identity.
In IoT deployments, such access can arise when an adversary extracts an on-device FR model from a physically compromised smart camera, access-control terminal, or mobile device through firmware inspection or exposed local storage. The extracted model can then be analyzed offline using external computing resources, so inversion need not run on the resource-constrained device itself. We therefore treat white-box inversion as a realistic high-access, worst-case privacy audit rather than a universal adversarial capability.

Despite the well-defined threat model, recovering high-fidelity images remains a formidable challenge due to the high dimensionality of the search space. Early approaches framed inversion as a direct optimization problem in the pixel space, aiming to maximize the likelihood of the target class. However, the optimization landscape of pixel values is highly non-convex, rendering these methods prone to entrapment in local minima. Consequently, the reconstructed images often suffer from pronounced visual artifacts and fail to preserve semantic coherence. To mitigate this, subsequent works leveraged Generative Adversarial Networks (GANs) \cite{goodfellow2020generative,gui2021review} as prior knowledge, optimizing a latent code within the generator's manifold. While GAN priors improve visual plausibility, they introduce a structural mismatch: the mapping from the latent space to the image space is inherently non-linear and non-smooth. This results in unstable gradient backpropagation, impeding the optimization from effectively traversing the manifold toward the precise features of the target identity.

In recent years, Diffusion Models (DMs)~\cite{lai_principles_2025,peebles2023scalable,song2023consistency} and their continuous-time generalizations, specifically Flow Matching (FM)~\cite{lipman_flow_2023}, have superseded GANs as the state-of-the-art in generative modeling. While these models offer superior capabilities in modeling complex data distributions, adapting them for adversarial inversion remains non-trivial. Most diffusion-based inversion methods rely on Stochastic Differential Equation (SDE)-based sampling. However, the stochastic noise injected along the SDE trajectory can mislead, or even interfere with, the sampling process toward the target class. Crucially, these methods lack a mathematically grounded mechanism to effectively rectify the generation path based on the target model's response. Consequently, they struggle to explicitly navigate the generative flow toward the specific identity regions defined by the target classifier.

In this paper, we bridge these gaps by proposing Steering Flow Model Inversion (SFMI), a Flow Matching-based white-box model inversion method that reformulates inversion as a trajectory-steering task. Unlike diffusion models that rely on complex stochastic differential equations, Flow Matching models data generation through an Ordinary Differential Equation (ODE), providing relatively smooth and stable velocity fields. We leverage this property to treat the inversion process not as a static optimization, but as guiding a flow from a simple prior distribution to the target class manifold. Specifically, SFMI operates in two stages: Step I, Learning a Generic Flow Matching Prior, which pre-trains an unconditional Flow Matching model to encapsulate a robust human face prior; and Step II, Attacking with Progressive Guidance Scheduler (PGS), which injects target-specific gradients during integration. By backpropagating through the target model at intermediate integration steps, PGS injects adaptive gradients into the velocity field, effectively ``steering'' the flow toward the high-density regions of the target identity.

We evaluate SFMI under an identity-disjoint protocol across diverse face recognition targets, including ResNet-based backbones, commonly used face recognition models such as ArcFace and CosFace, the modern transformer-based ViT model, and MobileFaceNet, which is widely adopted in IoT and edge-face-recognition scenarios.
SFMI achieves an average attack accuracy above 90\% with favorable perceptual similarity, demonstrating strong and competitive performance relative to state-of-the-art methods under the evaluated white-box protocol.

Our contributions can be summarized as follows: 
\begin{itemize}
  \item We propose SFMI, the first white-box model inversion method that leverages Flow Matching as the generative prior and formulates inversion as ODE velocity-field steering, thereby mitigating the optimization instability of previous GAN-based and stochastic diffusion-based attacks.
  \item We design the Progressive Guidance Scheduler (PGS) that seamlessly integrates the target model's feedback into the ODE solver. By dynamically modulating the vector field with adaptive gradient signals, we achieve fine-grained control over the identity recovery process. 
  \item Under an identity-disjoint cross-evaluation protocol, extensive experiments on diverse face recognition targets demonstrate that SFMI delivers strong and competitive performance relative to state-of-the-art optimization-based and generative MIA methods, particularly in attack accuracy and visual fidelity.
\end{itemize}

The remainder of this paper is organized as follows. Section~\ref{sec:related} reviews related work on optimization-based and generative-prior model inversion. Section~\ref{sec:method} introduces the proposed SFMI method, including \emph{Learning a Generic Flow Matching Prior} and \emph{Attacking with Progressive Guidance Scheduler (PGS)}. Section~\ref{sec:experiments} presents experimental settings and quantitative and qualitative evaluations. {Section~\ref{sec:ethics} discusses ethical considerations and responsible disclosure.} Finally, Section~\ref{sec:conclusion} concludes the paper and discusses future directions.

\section{Related Work}
\label{sec:related}

\subsection{Deep Face Recognition}
\label{subsec:deep_fr}

Deep learning architectures have substantially reshaped visual representation learning. Convolutional Neural Networks (CNNs) established a dominant paradigm for extracting hierarchical spatial features, Residual Networks (ResNet)~\cite{he_deep_2016} enabled stable training of deeper models, and Vision Transformers (ViT)~\cite{dosovitskiy_image_2021} further introduced self-attention for global contextual modeling.

Driven by these advances, face recognition has shifted from handcrafted-feature pipelines to deep learning-based paradigms~\cite{wang_deep_2020, du_elements_2021, wang_survey_2022, kim_50_2025}, where models learn identity-discriminative embeddings in an end-to-end manner. Training objectives have also evolved from metric-learning formulations, such as FaceNet~\cite{schroff_facenet_2015}, to margin-based softmax losses that enhance intra-class compactness and inter-class separability. In this work, we evaluate SFMI on representative targets spanning these developments: Face.evoLVe and IR-152 are conventional CNN-based FR backbones; CosFace~\cite{wang_cosface_2018} and ArcFace~\cite{deng_arcface_2022} represent widely used margin-based recognition objectives; MobileFaceNet~\cite{chen2018mobilefacenets} is a lightweight model designed for accurate real-time face verification on mobile, IoT, and edge devices; and the ViT model represents modern transformer-based FR designs. These heterogeneous targets provide different model capacities, embedding geometries, and deployment characteristics for evaluating inversion difficulty.

\subsection{Model Inversion Attacks}
\label{subsec:mia}

Model Inversion Attacks (MIAs) pose a severe privacy threat by aiming to reconstruct sensitive training data or infer private attributes from trained machine learning models \cite{fang_privacy_2024, yang_deep_2025}. The concept was initially formalized by exploiting prediction confidence information from simple classifiers to recover recognizable images \cite{fredrikson_model_2015}. Subsequent research rapidly expanded MIAs into rigorous white-box settings \cite{nasr_comprehensive_2019, song_machine_2017}, formulating the attack as an explicit optimization problem in continuous pixel space. By utilizing gradient descent \cite{yang_neural_2019}, attackers attempted to synthesize random noise into inputs that maximize the target class probability. However, because this optimization is performed directly in the extremely high-dimensional and highly non-convex natural image pixel space, traditional methods are prone to being trapped in poor local optima, often resulting in failure modes such as semantically meaningless reconstructions dominated by abstract high-frequency noise.

Related privacy and face-recognition studies in IoT contexts have investigated distinct but complementary directions, including secure face identification under fog computing \cite{hu2017security}, sparse low-rank representation for IoT face recognition \cite{yang2021sparse}, and practical feature inference attacks in vertical federated prediction \cite{yang2023practical}, collectively highlighting privacy-related risks in IoT scenarios.

To mitigate the mathematical ill-posedness of direct pixel optimization, a profound paradigm shift occurred with the introduction of auxiliary generative priors. Zhang et al. \cite{zhang_secret_2020} pioneered the Generative Model Inversion (GMI) attack, which leverages a pre-trained Generative Adversarial Network (GAN) to constrain the adversarial search space strictly to a natural image manifold. Following this breakthrough, a proliferation of generative MIAs emerged to enhance reconstruction fidelity and attack transferability. Notable advancements include modeling private data distributions (KEDMI) \cite{chen_knowledge-enriched_2021}, decoupling target models from priors via Plug \& Play Attacks (PPA) \cite{struppek_plug_2022}, and utilizing pseudo-label-guided conditional GANs (PLG) to disentangle search spaces \cite{yuan_pseudo_2023}. More recently, researchers have focused on dissecting the prior architecture itself by exploiting intermediate GAN features (IFGMI) \cite{qiu_closer_2024}. Diffusion-based MIA methods have also emerged as a recent direction: DiffMI~\cite{li2024model} fine-tunes a target-specific conditional diffusion prior to improve consistency with the target identity, while FGMIA~\cite{lu_fgmia_2025} first recovers feature encodings through a surrogate model and then performs one-shot feature-guided generation without iterative correction from the target model. Several other recent works have also explored related research directions~\cite{zhou2026model, shen2026swfti, wang2026gidd}.  Other explorations span variational inversion formulations \cite{wang_variational_2022}, optimization objective refinements \cite{nguyen_re-thinking_2023}, and adversarial prior exploitations \cite{usynin_beyond_2022, khosravy_model_2022}.

To clarify the methodological differences among representative MIAs, Table~\ref{tab:mia_related_comparison} summarizes their publication years, priors, evaluation data, image resolutions, attack knowledge, and gradient dependence.

\begin{table*}[t]
    \centering
    \caption{Comparison of representative model inversion attack methods in related work, including publication year, prior type, dataset, image resolution, attack scenario, and gradient requirement.}
    \label{tab:mia_related_comparison}
    \begin{tabularx}{0.95\textwidth}{l c X X c c c}
        \toprule
        Method & Year & Prior & Dataset & Resolution & Scenario & Gradients \\
        \midrule
        MIA~\cite{fredrikson_model_2015} & 2015 & None & MNIST & Various & Black/White-box & Optional \\
        YMIA~\cite{yang_neural_2019} & 2019 & None & AT\&T Faces & Various & Black/White-box & Optional \\
        GMI~\cite{zhang_secret_2020} & 2020 & Unconditional GAN & CelebA, FaceScrub & 64 & White-box & Yes \\
        KEDMI~\cite{chen_knowledge-enriched_2021} & 2021 & Knowledge-enriched GAN & CelebA, FaceScrub & 64 & White-box & Yes \\
        FMI~\cite{khosravy_model_2022} & 2022 & DCGAN & FaceScrub, CelebA & 64 & White-box & Yes \\
        PPA~\cite{struppek_plug_2022} & 2022 & StyleGAN2 & CelebA, FFHQ & 112/224 & White-box & Yes \\
        PLGMI~\cite{yuan_pseudo_2023} & 2023 & Pseudo-label cGAN & CelebA, FaceScrub, PubFig & 64/112 & White-box & Yes \\
        DiffMI~\cite{li2024model} & 2024 & Conditional DDPM & CelebA, FFHQ & 64/112 & White-box & Yes \\
        IFGMI~\cite{qiu_closer_2024} & 2024 & StyleGAN2 & CelebA, FaceScrub & 64/112/224 & White-box & Yes \\
        FGMIA~\cite{lu_fgmia_2025} & 2025 & DDPM & CelebA, FFHQ & 64/112/224 & Black/White-box & No \\
        \midrule
        SFMI (Ours) & 2026 & Flow Matching & CelebA, FFHQ & 64/112/224 & White-box & Yes \\
        \bottomrule
    \end{tabularx}
\end{table*}

As for defense methods, existing model-inversion defenses~\cite{fang_privacy_2024} mainly fall into two categories. The first category strengthens model training or pre-training, using robust representation learning, mutual-information regularization~\cite{wang2021improving}, transfer-learning-based robustness~\cite{ho2024model}, training-data deduplication~\cite{kandpal2022deduplicating}, trapdoor-based defense~\cite{liu2024trapmid}, or label smoothing~\cite{struppek2024careful} to reduce sensitive memorization. The second category processes model outputs, using noise injection, prediction purification~\cite{yang2020defending}, adversarial perturbation~\cite{wen2021defending}, or other post-processing strategies to weaken exploitable signals. For face recognition models, however, extracting identity-discriminative features is necessary for utility, while model inversion attacks exploit this feature extraction process, revealing the trade-off between recognition performance and privacy security.

Despite improving visual plausibility, existing generative MIAs still suffer from fundamental limitations when reconstructing identity-specific targets. On the one hand, GAN-based approaches rely on latent-space optimization, where the latent-to-image mapping is highly non-linear and often non-smooth, leading to unstable optimization dynamics and frequent convergence to suboptimal solutions. On the other hand, diffusion-based approaches typically adopt Stochastic Differential Equation (SDE)-based sampling, whose injected stochastic perturbations can continuously deflect the generation trajectory away from the target class signal. Although DiffMI introduces target-specific conditional diffusion fine-tuning, its DDPM-based sampling still faces potential trajectory deviation and requires multi-step cascaded backpropagation, which can mislead the generative prior during inversion. FGMIA reduces iterative optimization by using recovered feature encodings, but it does not use target-model feedback to progressively correct the generation trajectory. As a result, both paradigms struggle to consistently recover precise, identity-specific target face images, especially against modern margin-based FR models (e.g., CosFace and ArcFace in Sec.~\ref{subsec:deep_fr}).

This fundamental optimization instability directly motivates our proposed Steering Flow Model Inversion (SFMI). SFMI uses Flow Matching to construct a deterministic ODE trajectory and progressively injects target-model feedback during sampling. By reformulating the generative prior as a deterministic ODE with smooth trajectories and introducing a progressive guidance mechanism, SFMI provides a trajectory-steering backbone. This mechanism effectively bridges the semantic gap, enabling target-specific adversarial gradients to stably and precisely steer the generative trajectory toward the target identity and thereby overcome the limitations of both highly non-convex GANs and stochastic diffusion models.

\section{Method}
\label{sec:method}

\begin{table}[t]
    \centering
    \caption{Notation used for datasets and key variables.}
    \label{tab:notation_datasets}
    \footnotesize
    \begin{tabularx}{\columnwidth}{lX}
        \toprule
        \textbf{Notation} & \textbf{Description} \\
        \midrule
        $\mathcal{D}_{\text{priv}}$ & Private dataset used to train the target face recognition model. \\
        $\mathcal{D}_{\text{pub}}$ & Public auxiliary dataset available to the adversary and strictly identity-disjoint from $\mathcal{D}_{\text{priv}}$. \\
        $y_{\text{target}}$ & Target identity label to be reconstructed by the attack. \\
        $x_0$ & Initial Gaussian noise sample drawn from $p_0=\mathcal{N}(0,\mathbf{I})$. \\
        $x_1$ & Clean face sample or the final endpoint of the generative trajectory. \\
        $x_t$ & Intermediate state along the flow trajectory at time $t\in[0,1]$. \\
        $u_t(\cdot)$ & Ground-truth conditional vector field defined by the OT path. \\
        $v_{\phi}(\cdot)$ & Learned velocity field induced by the Flow Matching prior $\mathcal{M}_{\phi}$. \\
        $\mathcal{M}_{\phi}$ & Flow Matching prior model parameterized by $\phi$. \\
        $\mathcal{L}_{\text{FM}}$ & Flow Matching training loss for learning $v_{\phi}$. \\
        $\mathcal{L}_{\text{id}}$ & Identity supervision objective used to guide inversion toward $y_{\text{target}}$. \\
        $g(x_t)$ & Raw identity-gradient signal computed with respect to the current state $x_t$. \\
        $g_{\mathrm{norm}}(x_t)$ & Normalized guidance signal scaled to the velocity magnitude. \\
        $\mathbb{E}$ & Expectation over sampled time steps, noise samples, and public data samples. \\
        \bottomrule
    \end{tabularx}
\end{table}

\subsection{Preliminaries}
\label{subsec:preliminaries}

\textbf{Problem Definition.} Let $f_{\theta}: \mathcal{X} \rightarrow \mathcal{Y}$ denote a pre-trained target face recognition model parameterized by $\theta$, which maps the continuous image space $\mathcal{X} \subseteq \mathbb{R}^{C \times H \times W}$ to the probability simplex $\mathcal{Y}$ over $K$ identities. Given a target label $y_{\text{target}} \in \{1, \dots, K\}$, the goal of a Model Inversion Attack (MIA) is to recover an image that captures the private visual characteristics of that identity from the information encoded in the trained model. We formulate this objective as an optimization problem over $\mathcal{X}$: the adversary seeks an input $x^*$ whose prediction is aligned with $y_{\text{target}}$, while a regularization term enforces natural image priors. Mathematically, the adversary solves the following minimization problem:
\begin{equation}
x^* = \mathop{\arg\min}_{x \in \mathcal{X}} \; \mathcal{L}_{\text{inv}}\bigl(f_{\theta}(x), y_{\text{target}}\bigr) + \lambda \mathcal{R}(x),
\label{eq:inversion_obj}
\end{equation}
where $\mathcal{L}_{\text{inv}}$ represents the classification objective that penalizes deviations from the target class, $\mathcal{R}(x)$ serves as a prior constraint to ensure the semantic plausibility of the reconstructed image, and $\lambda$ is a hyperparameter balancing the two terms.
In SFMI, $\mathcal{R}(x)$ is not evaluated or optimized as an explicit regularization loss. Instead, the Flow Matching model trained on public face data implicitly imposes the facial prior, while progressive target-model gradients steer the sampling trajectory toward the target class within the learned face manifold.

\textbf{Adversary Knowledge.} We operate under a white-box threat model, which grants the adversary comprehensive access to the target system. In this setting, the adversary is assumed to possess full knowledge of the model architecture and its trained parameters $\theta$, thereby enabling the explicit computation of gradients via backpropagation. Furthermore, to facilitate the generation of realistic facial structures, the adversary is assumed to have access to a generic public face dataset $\mathcal{D}_{\text{pub}}$ that remains strictly identity-disjoint from the private training set $\mathcal{D}_{\text{priv}}$ used for $f_{\theta}$. In practical terms, the public dataset shares neither identities nor samples with the private data. This independent data separation allows the adversary to learn transferable facial priors while fully preserving the assumption that private data are inaccessible.

\subsection{Steering Flow Model Inversion (SFMI)} \label{subsec:sfmi_overview}

\subsubsection{Method Overview} \label{subsubsec:overview}

\begin{figure*}[t]
  \centering
  \includegraphics[width=0.95\linewidth]{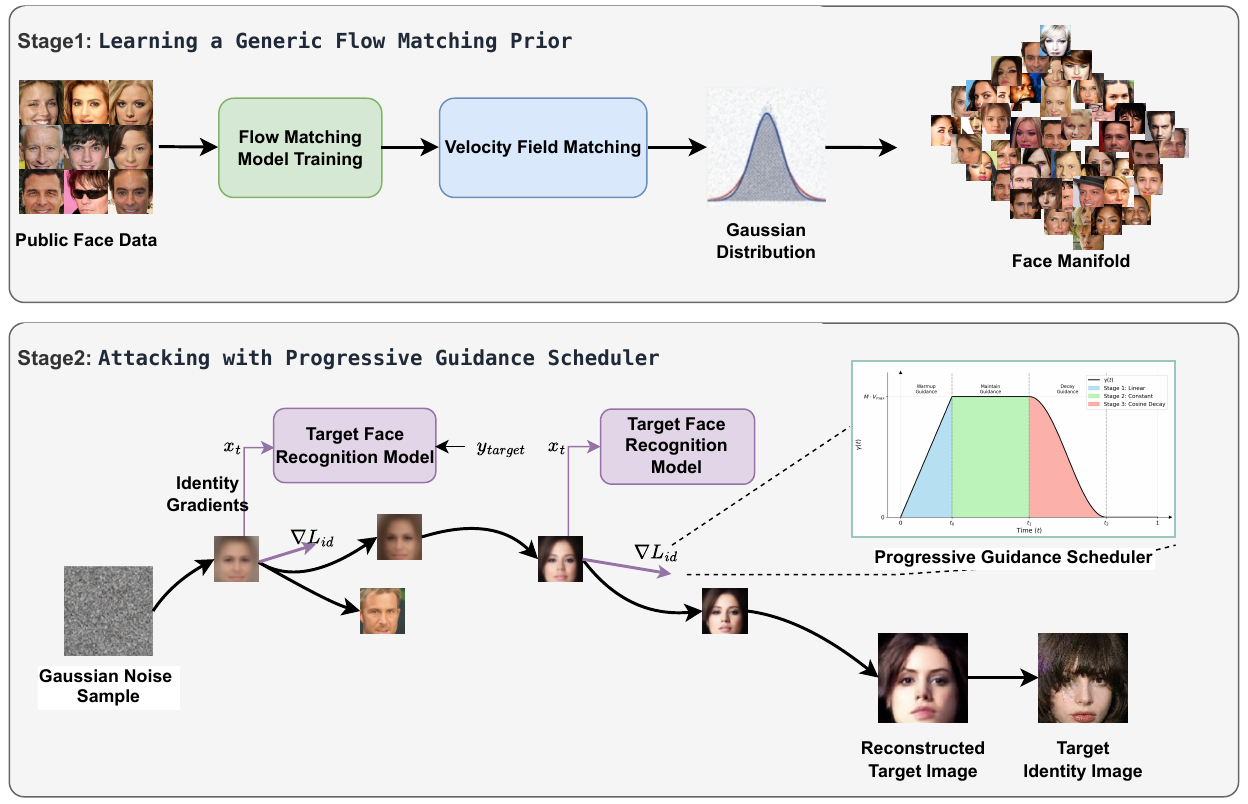}
  \caption{Overview of SFMI, which consists of two stages: learning a generic Flow Matching prior and performing progressive gradient-guided attack.}
  \label{fig:framework}
\end{figure*}

In this section, we first provide an overview of Steering Flow Model Inversion (SFMI). As illustrated in Fig.~\ref{fig:framework}, the method consists of two tightly coupled stages: Stage I provides a smooth generative prior over human faces, while Stage II injects target-aware supervision to steer this prior toward identity-specific reconstruction.

In the upper part, Stage I trains a Flow Matching model through velocity-field matching so that samples from a Gaussian prior can be transported to the human-face manifold. Specifically, the model learns a smooth vector field that maps samples from $p_0$ to samples from $p_1$, yielding a stable transport trajectory and a generic facial prior for inversion.
This prior captures broad facial structure and appearance statistics, which helps constrain subsequent attack trajectories to remain on realistic face manifolds.

In the lower part, Stage II starts from Gaussian noise and performs guided sampling toward a target identity. During sampling, clean estimates predicted from intermediate states are fed into the target model to compute identity-related gradients, and these gradients are injected back into the update direction to progressively move intermediate states toward the target face class.
Meanwhile, PGS dynamically modulates the guidance strength over time (e.g., warm-up, sustain, and gradual decay), balancing identity alignment and visual fidelity, and thus enabling stable trajectory correction throughout generation.

\subsubsection{Learning a Generic Flow Matching Prior}
\label{subsubsec:flow_matching_prior}

In this stage, we adopt Flow Matching (FM~\cite{lipman_flow_2023, esser_scaling_2024}) because model inversion requires a generative prior that is both smooth and controllable when transporting samples from noise to realistic face images. FM directly learns a time-dependent vector field, which provides a stable and differentiable trajectory for optimization and avoids the instability introduced by highly irregular mappings or stochastic perturbations. The role of this stage is therefore to learn an unconditional facial prior that maps a simple Gaussian distribution to the human-face manifold, providing a reliable generative backbone for subsequent identity-specific steering.

Let $\mathcal{D}_{\text{pub}}$ denote a public face dataset accessible to the adversary, which is disjoint from the private dataset $\mathcal{D}_{\text{priv}}$ used to train the target model. We define $p_1(x)$ as the empirical data distribution supported by $\mathcal{D}_{\text{pub}}$. Conversely, let $p_0(x)$ denote a simple prior noise distribution, which we specify as a standard $d$-dimensional Gaussian distribution, i.e., $p_0 = \mathcal{N}(0, \mathbf{I})$. Our objective is to learn a time-dependent vector field that pushes samples from $p_0$ to $p_1$.

We define this generative process via an Optimal Transport (OT~\cite{lipman_flow_2023}) displacement map. This map constructs the simplest possible trajectory---a straight line---between a noise sample $x_0 \sim p_0$ and a data sample $x_1 \sim p_1$. The state $x_t$ at any time $t \in [0, 1]$ is given by linear interpolation:
\begin{equation}
  x_t = \psi_t(x_0, x_1) = (1 - t)x_0 + t x_1.
  \label{eq:ot_path}
\end{equation}
Differentiating Eq.~\eqref{eq:ot_path} with respect to time yields the ground-truth conditional vector field, denoted as $u_t$, which represents the target velocity for this specific path:
\begin{equation}
  u_t(x\mid x_0, x_1) = x_1 - x_0.
  \label{eq:ot_target_velocity}
\end{equation}

While standard Flow Matching directly approximates $u_t$ with a velocity-predicting network, we opt for an $x$-prediction parameterization following prior practice~\cite{li_back_2025}. We train a neural network $\mathcal{M}_{\phi}(x_t, t)$ to estimate the clean original image $\hat{x}_1$ from the noisy state $x_t$. Based on the geometry of the OT path, the relationship between the current state $x_t$, the destination $x_1$, and the velocity $v$ is derived as $v = (x_1 - x_t) / (1 - t)$. Consequently, the vector field $v_{\phi}$ induced by our network is formulated as:
\begin{equation}
  v_{\phi}(x_t, t) = \frac{\mathcal{M}_{\phi}(x_t, t) - x_t}{1 - t}.
  \label{eq:induced_velocity}
\end{equation}

Following the design choice in~\cite{li_back_2025}, we optimize the model with a velocity-matching loss ($v$-loss). To prioritize training on the most critical temporal regions, we sample the time step $t$ from a Logit-Normal distribution parameterized by mean $m$ and standard deviation $\sigma$, rather than a uniform distribution. The loss is computed by minimizing the discrepancy between the induced velocity field $v_{\phi}$ (Eq.~\ref{eq:induced_velocity}) and the ground-truth target velocity $u_t$ (Eq.~\ref{eq:ot_target_velocity}):
\begin{equation}
  \mathcal{L}_{\text{FM}}(\phi) = \mathbb{E}_{t, x_0, x_1} \left[ \left\| v_{\phi}(x_t, t) - u_t \right\|_2^2 \right].
  \label{eq:fm_loss}
\end{equation}

The complete training process for the generic Flow Matching prior is detailed in Algorithm~\ref{alg:training}. At each iteration, we first sample a mini-batch of clean faces $x_1 \sim \mathcal{D}_{\text{pub}}$ and a mini-batch of Gaussian noise samples $x_0 \sim \mathcal{N}(0, \mathbf{I})$, then draw time steps $t$ from a Logit-Normal distribution with $\text{logit}(t) \sim \mathcal{N}(m, \sigma^2)$ to emphasize informative temporal regions. Next, we construct intermediate states along the OT path via linear interpolation and compute the corresponding target velocities $u_t = x_1 - x_0$. Given each pair $(x_t, t)$, the network $\mathcal{M}_{\phi}$ predicts $\hat{x}_1$, which is converted to the induced velocity $v_{\phi}$ using Eq.~\ref{eq:induced_velocity}. We then evaluate the $v$-loss in Eq.~\ref{eq:fm_loss} as the batch-wise discrepancy between $v_{\phi}$ and $u_t$, and update parameters $\phi$ by gradient descent. By repeating this procedure, the model learns a stable, smooth vector field that reliably transports Gaussian noise toward realistic face samples.

\begin{algorithm}[t]
\caption{Training of the Generic Flow Matching Prior}
\label{alg:training}
\begin{algorithmic}[1]
\renewcommand{\algorithmicrequire}{\textbf{Input:}}
\renewcommand{\algorithmicensure}{\textbf{Output:}}
\REQUIRE Public dataset $\mathcal{D}_{\text{pub}}$, max iterations $N$, batch size $B$, learning rate $\eta$, sampling hyperparameters $m, \sigma$.
\ENSURE Trained model $\mathcal{M}_{\phi}$.
\STATE Initialize network parameters $\phi$.
\FOR{$iter = 1$ to $N$}
  \STATE Sample data batch $\{x_1^{(i)}\}_{i=1}^B \sim \mathcal{D}_{\text{pub}}$.
  \STATE Sample noise batch $\{x_0^{(i)}\}_{i=1}^B \sim \mathcal{N}(0, \mathbf{I})$.
  \STATE Sample time steps $\{t^{(i)}\}_{i=1}^B \sim \text{Logit-Normal}(m, \sigma)$. \hfill \COMMENT{$\text{logit}(t) \sim \mathcal{N}(m, \sigma^2)$}
  \FOR{$i = 1$ to $B$}
    \STATE Construct intermediate state via OT path:
    \STATE $x_t^{(i)} \leftarrow (1 - t^{(i)})x_0^{(i)} + t^{(i)} x_1^{(i)}$
    \STATE Forward pass to predict clean data:
    \STATE $\hat{x}_1^{(i)} \leftarrow \mathcal{M}_{\phi}(x_t^{(i)}, t^{(i)})$
    \STATE Compute induced velocity (Eq.~\ref{eq:induced_velocity}):
    \STATE $v_{\phi}^{(i)} \leftarrow (\hat{x}_1^{(i)} - x_t^{(i)}) / (1 - t^{(i)})$
    \STATE Compute target velocity (Eq.~\ref{eq:ot_target_velocity}):
    \STATE $u_t^{(i)} \leftarrow x_1^{(i)} - x_0^{(i)}$
  \ENDFOR
  \STATE Compute Loss: $\mathcal{L} \leftarrow \frac{1}{B} \sum_{i=1}^B \| v_{\phi}^{(i)} - u_t^{(i)} \|_2^2$
  \STATE Update parameters: $\phi \leftarrow \phi - \eta \nabla_{\phi} \mathcal{L}$
\ENDFOR
\RETURN Trained model $\mathcal{M}_{\phi}$.
\end{algorithmic}
\end{algorithm}

\subsubsection{Progressive Gradient-Guided Attack}
\label{subsubsec:guided_attack}

With the generic flow matching prior $\mathcal{M}_{\phi}$ trained, the second stage of SFMI focuses on recovering the specific target identity. We formulate this as a trajectory steering problem within the ODE formulation. The objective is to navigate the generative flow such that the final sample minimizes the identity mismatch with respect to the target label $y_{\text{target}}$.

To achieve this, we intervene in the numerical integration process. In a standard unconditional generation, the trajectory is governed solely by the learned velocity field $v_{\phi}(x_t, t)$. In our adversarial setting, we introduce a guidance term derived from the target classifier $f_{\theta}$. Crucially, since the target classifier is trained on clean images, directly feeding the noisy state $x_t$ into $f_{\theta}$ would yield uninformative gradients. Instead, we leverage the $x$-prediction capability of our pre-trained prior.

At each time step $t$, we first pass the current noisy state $x_t$ through the Flow Matching model to obtain a clean image estimate $\hat{x}_1 = \mathcal{M}_{\phi}(x_t, t)$. This estimated clean image is then fed into the target classifier to compute the inversion loss. Formally, the loss function is defined as the composition of the classifier and the prior model:
\begin{equation}
  \mathcal{L}_{\text{inv}}(x_t) = \mathcal{L}_{\text{id}}\bigl(f_{\theta}(\mathcal{M}_{\phi}(x_t, t)), y_{\text{target}}\bigr).
  \label{eq:attack_loss}
\end{equation}
where $\mathcal{L}_{\text{id}}$ denotes the identity supervision objective. In practice, we instantiate $\mathcal{L}_{\text{id}}$ with the max-margin loss (MMLoss) adopted by PLGMI~\cite{yuan_pseudo_2023}. Let $z_{\theta}(\hat{x}_1) \in \mathbb{R}^{K}$ be the logit vector of the target classifier for the estimated clean image $\hat{x}_1=\mathcal{M}_{\phi}(x_t,t)$. MMLoss is defined as
\begin{equation}
  \mathcal{L}_{\text{MM}}(z_{\theta}(\hat{x}_1), y_{\text{target}})
  = \max_{k \ne y_{\text{target}}} z_{\theta,k}(\hat{x}_1)
  - z_{\theta,y_{\text{target}}}(\hat{x}_1).
\label{eq:mm_loss}
\end{equation}
Minimizing Eq.~\ref{eq:mm_loss} enlarges the target logit relative to the strongest non-target logit, which encourages identity-discriminative reconstruction.

To steer the flow, we require the gradient with respect to the current state $x_t$, not the estimated outcome $\hat{x}_1$. We obtain this by backpropagating the error signal through the frozen target classifier and, significantly, through the frozen Flow Matching prior itself:
\begin{equation}
  g(x_t) = \nabla_{x_t} \mathcal{L}_{\text{inv}}(x_t)
  = \nabla_{x_t} \left[ \mathcal{L}_{\text{id}}\bigl(f_{\theta}(\mathcal{M}_{\phi}(x_t, t)), y_{\text{target}}\bigr) \right].
  \label{eq:gradient_xt_chain}
\end{equation}
This end-to-end backpropagation ensures that the guidance signal $g(x_t)$ accurately reflects how an infinitesimal change in the current noisy state $x_t$ influences the final identity objective, accounting for the manifold projection learned by $\mathcal{M}_{\phi}$. Before injection, we normalize the raw gradient to a velocity-proportional length:
\begin{equation}
  g_{\mathrm{norm}}(x_t)
  = \frac{g(x_t)}{\|g(x_t)\|_2 + \epsilon} \cdot \|v_{\phi}(x_t,t)\|_2,
\label{eq:normalized_guidance}
\end{equation}
where $\epsilon$ is a small constant for numerical stability. This preserves the adversarial direction while matching its scale to the FM velocity.

We then modify the original velocity field by injecting this gradient signal. The rectified velocity field $\tilde{v}(x_t, t)$ is formulated as:
\begin{equation}
  \tilde{v}(x_t, t) = v_{\phi}(x_t, t) - \gamma(t) \cdot g_{\mathrm{norm}}(x_t).
  \label{eq:steered_velocity}
\end{equation}
As illustrated in Fig.~\ref{fig:framework}, this correction preserves the natural FM trajectory while steering sampling toward a face recognized as the target identity.

\textbf{Progressive Guidance Schedule.} Instead of a constant guidance scale, which often leads to saturation artifacts or optimization instability, we design a time-dependent schedule $\gamma(t)$ that modulates the steering strength dynamically. The schedule consists of four phases: warm-up, sustain, decay, and relaxation. Mathematically, it is defined as:
\begin{equation}
  \gamma(t) = M \cdot
  \begin{cases}
    V_{\max} \cdot \frac{t}{t_0}, & 0 \le t \le t_0, \\
    V_{\max}, & t_0 < t \le t_1, \\
    \frac{V_{\max}}{2} \left[ 1 + \cos\left( \pi \frac{t - t_1}{t_2 - t_1} \right) \right], & t_1 < t \le t_2, \\
    0, & t_2 < t \le 1.
  \end{cases}
  \label{eq:guidance_schedule}
\end{equation}
where $M$ represents the global magnitude scale, and $V_{\max}$ defines the peak intensity. The thresholds $t_0, t_1, t_2$ delineate the phases. The linear warm-up ($t \le t_0$) prevents abrupt trajectory shifts when noise is high. The cosine decay ($t_1 < t \le t_2$) ensures a smooth transition, and the final zero-guidance phase ($t > t_2$) allows the Flow Matching prior to refine the image without external interference, thereby reinforcing adherence to the learned face manifold, stabilizing the final refinement trajectory, and preserving high visual fidelity.
The PGS diagram in Fig.~\ref{fig:framework} visualizes this schedule as four consecutive warm-up, sustain, decay, and relaxation stages, providing an intuitive counterpart to the piecewise definition above.

To mitigate discretization errors and ensure the steered trajectory adheres smoothly to the face manifold, we adopt a second-order predictor-corrector scheme inspired by the Heun sampler~\cite{Karras2022edm}. As detailed in Algorithm~\ref{alg:attack}, this mechanism involves a two-stage evaluation at each step: a predictor step to estimate a provisional next state using the gradient at the current position, and a corrector step to refine the update using the gradient at the predicted position. This approach effectively corrects the trajectory curvature induced by the adversarial guidance. Following common practice, we omit the final Heun corrector and use the predictor output directly at the last step.

\begin{algorithm}[t]
\caption{SFMI: Progressive Gradient-Guided Attack}
\label{alg:attack}
\begin{algorithmic}[1]
\renewcommand{\algorithmicrequire}{\textbf{Input:}}
\renewcommand{\algorithmicensure}{\textbf{Output:}}
\REQUIRE Target model $f_{\theta}$, Prior model $\mathcal{M}_{\phi}$, Target label $y_{\text{target}}$, Steps $N$.

\ENSURE Reconstructed image $x_1$.
\STATE Sample initial noise $x_0 \sim \mathcal{N}(0, \mathbf{I})$.
\STATE Set initial state $x_{t_0} \leftarrow x_0$ and step size $\Delta t = 1/N$.
\FOR{$i = 0$ to $N-1$}
  \STATE $t_i \leftarrow i/N$; \quad $t_{i+1} \leftarrow (i+1)/N$.
  \STATE Compute $\gamma(t_i)$ and $\gamma(t_{i+1})$ using Eq.~\ref{eq:guidance_schedule}.
  \STATE \textit{// Stage 1: Predictor}
  \STATE $\hat{x}_1 \leftarrow \mathcal{M}_{\phi}(x_{t_i}, t_i)$.
  \STATE $v_{\text{curr}} \leftarrow (\hat{x}_1 - x_{t_i}) / (1 - t_i)$.
  \STATE $g_{\text{curr}} \leftarrow \nabla_{x_{t_i}} \mathcal{L}_{\text{id}}\bigl(f_{\theta}(\mathcal{M}_{\phi}(x_{t_i}, t_i)), y_{\text{target}}\bigr)$.
  \STATE $g_{\text{curr}} \leftarrow \frac{g_{\text{curr}}}{\|g_{\text{curr}}\|_2+\epsilon}\cdot\|v_{\text{curr}}\|_2$.
  \STATE $\tilde{v}_1 \leftarrow v_{\text{curr}} - \gamma(t_i) \cdot g_{\text{curr}}$.
  \STATE $\hat{x}_{t_{i+1}} \leftarrow x_{t_i} + \tilde{v}_1 \cdot \Delta t$.

  \IF{$i=N-1$}
    \RETURN $x_1 = \hat{x}_{t_{i+1}}$.
  \ENDIF

  \STATE \textit{// Stage 2: Corrector}
  \STATE $\hat{x}'_1 \leftarrow \mathcal{M}_{\phi}(\hat{x}_{t_{i+1}}, t_{i+1})$.
  \STATE $v_{\text{pred}} \leftarrow (\hat{x}'_1 - \hat{x}_{t_{i+1}}) / (1 - t_{i+1})$.
  \STATE $g_{\text{pred}} \leftarrow \nabla_{\hat{x}_{t_{i+1}}} \mathcal{L}_{\text{id}}\bigl(f_{\theta}(\mathcal{M}_{\phi}(\hat{x}_{t_{i+1}}, t_{i+1})), y_{\text{target}}\bigr)$.
  \STATE $g_{\text{pred}} \leftarrow \frac{g_{\text{pred}}}{\|g_{\text{pred}}\|_2+\epsilon}\cdot\|v_{\text{pred}}\|_2$.
  \STATE $\tilde{v}_2 \leftarrow v_{\text{pred}} - \gamma(t_{i+1}) \cdot g_{\text{pred}}$.

  \STATE \textit{// Update State}
  \STATE $x_{t_{i+1}} \leftarrow x_{t_i} + \frac{\Delta t}{2}(\tilde{v}_1 + \tilde{v}_2)$.
\ENDFOR
\RETURN $x_1 = x_{t_N}$.
\end{algorithmic}
\end{algorithm}

\section{Experiments}\label{sec:experiments}
{This section presents the experimental protocol, comparisons with representative white-box inversion baselines, component ablations, and an analysis of guidance variants and hyperparameter sensitivity.}
{A separate defense evaluation examines output perturbation and training-phase regularization.}

\subsection{Experimental Setup}\label{subsec:exp_setup}
\textbf{Datasets.} We enforce an identity-disjoint protocol throughout all experiments: identities in the adversary's public prior data never overlap with those used to train the target models. Consistent with Sec.~\ref{sec:method}, $\mathcal{D}_{\text{priv}}$ denotes the private target-training dataset, and $\mathcal{D}_{\text{pub}}$ denotes the public auxiliary dataset available to the adversary. From CelebA~\cite{liu_deep_2015}, we construct CelebA-priv as $\mathcal{D}_{\text{priv}}$ with 1000 identities and 30{,}000 images for target-model training, and build a non-overlapping 30{,}000-image CelebA-pub split as the default $\mathcal{D}_{\text{pub}}$ for learning the unconditional face prior. Since prior learning is unconditional, $\mathcal{D}_{\text{pub}}$ requires no identity annotations. To evaluate cross-distribution transferability, we also construct FFHQ-pub from FFHQ~\cite{karras2019style} as another identity-disjoint 30{,}000-image auxiliary prior dataset. Thus, CelebA-pub and FFHQ-pub use the same public-data scale while preventing overlap with CelebA-priv. All images follow the same face-recognition preprocessing pipeline, including face detection, five-point landmark localization, 2D partial affine alignment, interpolation, cropping, and resizing to the target-model resolution.

\textbf{Target Models.} We evaluate SFMI against six face recognition targets covering different architectures, objectives, and input resolutions. Face.evoLVe~\cite{8265437} and IR-152~\cite{he_deep_2016} are conventional CNN-based face recognition models evaluated at $64\times64$ resolution. CosFace~\cite{wang_cosface_2018} and ArcFace~\cite{deng_arcface_2022} are two widely used margin-based face recognition models and are evaluated at $112\times112$ resolution. MobileFaceNet~\cite{chen2018mobilefacenets} is a lightweight face recognition architecture commonly used in mobile, IoT, and edge-device scenarios, and ViT~\cite{dosovitskiy_image_2021} represents a modern transformer-based face recognition model; both are evaluated at $112\times112$ resolution in the main comparison. These resolutions refer to the detected, aligned, cropped, and resized face inputs consumed by the FR models rather than raw high-resolution images. In the ablation study, we further include $224\times224$ ArcFace and MobileFaceNet targets to examine higher-resolution behavior. For each target, we use an Internet-public pretrained model as the backbone and train only the final classification layer on $\mathcal{D}_{\text{priv}}$ for 10 epochs. To ensure a rigorous and unbiased assessment of the recovered identities and avoid self-evaluation, we strictly adopt a cross-evaluation protocol, where the quantitative performance of an attack against a specific target model is reported as the average score evaluated by the other models.

\textbf{Evaluation Metrics.} We evaluate model inversion performance using three complementary metrics: Attack Accuracy (\textbf{ACC}), Fr\'echet Inception Distance (\textbf{FID}), and Learned Perceptual Image Patch Similarity (\textbf{LPIPS}). Together, they characterize identity recovery success, global visual realism, and fine-grained perceptual consistency. ACC quantifies semantic identity recovery under the cross-evaluation protocol by computing the average proportion of reconstructed images that are recognized as the target identity by non-target evaluators, thus directly reflecting attack effectiveness. During the attack, the adversary only accesses the target model and has no access to the evaluation model; therefore, ACC measures whether the reconstruction is recognized as the target identity by independent evaluators rather than merely fitting the attacked model. FID~\cite{heusel2017gans} measures distribution-level similarity and overall image fidelity via the Wasserstein-2 distance between reconstructed samples and real private training images in feature space, where lower values indicate fewer artifacts and better photorealism. LPIPS~\cite{zhang2018unreasonable} further evaluates structural and textural alignment between reconstructions and ground-truth faces using deep perceptual features, and lower LPIPS indicates closer perceptual resemblance to the target identity.

\textbf{Implementation Details.} We implement $\mathcal{M}_{\phi}$ using the Flow Matching architecture in~\cite{li_back_2025}, which is a ViT/DiT-like Transformer with a $16\times16$ patch size and 16 attention heads. The FM prior is trained for 500k optimization steps using AdamW, a per-batch learning rate of $2\times10^{-5}$, and a learning-rate schedule with linear warmup followed by cosine decay. Instead of uniformly sampling $t$, we use $\mathrm{logit}(t)\sim\mathcal{N}(-0.8,0.8^2)$. During the attack, we use 50 sampling steps by default, with $M=0.3$, $t_0=0.1$, $t_1=0.3$, $t_2=0.7$, and $V_{\max}=1.0$. To balance the facial prior and identity guidance, SFMI directly predicts the original image with the FM prior, injects normalized guidance only through the selected PGS stages, and strictly controls the guidance magnitude; in practice, we recommend keeping $M\le 0.35$ because larger values can push the trajectory too far and cause facial distortion or artifacts. At the $112\times112$ resolution setting, training one FM prior takes about 18 hours on a single RTX 5090 GPU with 32 GB of memory; at the $224\times224$ resolution setting, it takes about 50 hours on the same GPU. Once trained, the generic FM prior can be reused for different target models. The 50-step attack takes about 0.23 seconds per image at $112\times112$ resolution and 0.47 seconds per image at $224\times224$ resolution.

\subsection{Comparison with State-of-the-Art Methods}\label{subsec:sota_comparison}
To verify the effectiveness of SFMI, we conduct a comprehensive comparison with existing white-box model inversion methods. Specifically, we evaluate six target face recognition models (Face.evoLVe, IR-152, CosFace, ArcFace, MobileFaceNet, and ViT), following prior settings~\cite{yuan_pseudo_2023, lu_fgmia_2025} to ensure consistency with standard white-box MIA evaluation practice. We compare SFMI against five representative baselines: GMI~\cite{zhang_secret_2020}, PPA~\cite{struppek_plug_2022}, PLGMI~\cite{yuan_pseudo_2023}, IFGMI~\cite{qiu_closer_2024}, and FGMIA~\cite{lu_fgmia_2025}. For fairness, all methods are implemented from official codebases and evaluated under the same protocol in Sec.~\ref{subsec:exp_setup}, including strict identity-disjoint data splits, model-aligned preprocessing, 30{,}000-image public auxiliary data at comparable scale, and the same cross-evaluation rule. Specifically, SFMI uses CelebA-pub as the default $\mathcal{D}_{\text{pub}}$ for training the Flow Matching prior. For the baselines, GMI, PLGMI, and FGMIA use priors trained following their original procedures with the same-scale public auxiliary data, while PPA and IFGMI reuse the prior models released by the original papers because their official protocols rely on these pretrained generative priors. All baseline attacks use the parameters recommended by the corresponding papers or official implementations.

\begin{table*}[t]
    \centering
    \caption{Computational costs for $112\times112$ CosFace inversion on an RTX 5090 (32\,GB). Prior-training hardware is listed separately; memory denotes device capacity. See Sec.~\ref{subsec:exp_setup} for SFMI timings at $224\times224$.}
    \label{tab:baseline_config}
    \footnotesize
    \setlength{\tabcolsep}{3pt}
    \renewcommand{\arraystretch}{1.08}
    \begin{tabularx}{\textwidth}{l l c c >{\raggedright\arraybackslash}X c}
        \toprule
        \textbf{Method} & \textbf{Prior type} & \textbf{Prior-training device} & \textbf{Prior-training time} & \textbf{Attack budget} & \textbf{Inversion time} \\
        \midrule
        GMI~\cite{zhang_secret_2020} & WGAN-GP~\cite{gulrajani2017improved} & RTX 5090 (32\,GB) & 15 h & 1500 updates & 0.16 s/pic \\
        PPA~\cite{struppek_plug_2022} & StyleGAN2~\cite{karras2020analyzing} & $8\times$ Tesla V100 (16\,GB) & 234 h (pretrained) & 5K candidates + 70 updates & 0.42 s/pic \\
        PLGMI~\cite{yuan_pseudo_2023} & Pseudo-label cGAN & RTX 5090 (32\,GB) & 17 h & 600 updates & 0.20 s/pic \\
        IFGMI~\cite{qiu_closer_2024} & StyleGAN2~\cite{karras2020analyzing} & $8\times$ Tesla V100 (16\,GB) & 234 h (pretrained) & 5K candidates + 170 updates & 0.27 s/pic \\
        FGMIA~\cite{lu_fgmia_2025} & Feature-guided DDPM & RTX 5090 (32\,GB) & 13 h & 50 denoising steps & 0.40 s/pic \\
        SFMI (Ours) & Flow Matching & RTX 5090 (32\,GB) & 18 h & 50 steps (99 NFE) & 0.23 s/pic \\
        \bottomrule
    \end{tabularx}
\end{table*}

Table~\ref{tab:baseline_config} jointly summarizes the prior architecture, per-attack computational budget, prior-training time, and attack time. PPA and IFGMI use released StyleGAN2 checkpoints rather than priors retrained on our platform. For context, the official StyleGAN2 implementation reports 9 days and 18 hours to train its FFHQ model at $1024\times1024$ resolution on eight Tesla V100 GPUs.
{These upstream training costs use a different hardware and resolution setting, while all inversion times are measured on a single RTX 5090 GPU with 32\,GB memory.}
For locally measured training and inversion times, each method uses the largest batch size that fits within the RTX 5090's 32\,GB capacity. For the eight-V100 StyleGAN2 training configuration, the measured peak training memory is approximately 14.5\,GB per GPU.
With the final Heun corrector omitted, SFMI uses $2N-1$ prior evaluations, corresponding to 99 NFE for its 50-step attack. For SFMI, we repeat the full process three times, from Flow Matching prior training to attacking each target class; for each target class, eight images are sampled and used to compute the reported mean and standard deviation.

\begin{table*}[t]
    \centering
    \caption{Comprehensive white-box MIA comparison among GMI~\cite{zhang_secret_2020}, PPA~\cite{struppek_plug_2022}, PLGMI~\cite{yuan_pseudo_2023}, IFGMI~\cite{qiu_closer_2024}, FGMIA~\cite{lu_fgmia_2025}, and SFMI on six target models (Face.evoLVe, IR-152, CosFace, ArcFace, MobileFaceNet, and ViT). Metrics include ACC ($\uparrow$), FID ($\downarrow$), and LPIPS ($\downarrow$) under the cross-evaluation protocol. Best results are highlighted in \textbf{bold}. All methods are reported as mean $\pm$ standard deviation over three repeated runs.}
    \label{tab:quantitative_sota}
    {\footnotesize
    \setlength{\tabcolsep}{4pt}
    \renewcommand{\arraystretch}{1.08}
        \resizebox{\textwidth}{!}{\begin{tabular}{l|ccc|ccc|ccc}
            \toprule
            \multirow{2}{*}{\textbf{Method}} & \multicolumn{3}{c|}{\textbf{Target: Face.evoLVe ($64\times64$)}} & \multicolumn{3}{c|}{\textbf{Target: IR-152 ($64\times64$)}} & \multicolumn{3}{c}{\textbf{Target: CosFace ($112\times112$)}} \\
            \cmidrule{2-10}
            & \textbf{ACC} $\uparrow$ & \textbf{FID} $\downarrow$ & \textbf{LPIPS} $\downarrow$ & \textbf{ACC} $\uparrow$ & \textbf{FID} $\downarrow$ & \textbf{LPIPS} $\downarrow$ & \textbf{ACC} $\uparrow$ & \textbf{FID} $\downarrow$ & \textbf{LPIPS} $\downarrow$ \\
            \midrule
            GMI~\cite{zhang_secret_2020}     & $0.2537\pm\mathrm{0.0156}$ & $27.97\pm\mathrm{0.90}$ & $0.5319\pm\mathrm{0.0087}$ & $0.1382\pm\mathrm{0.0176}$ & $\mathbf{27.61}\pm\mathrm{0.40}$ & $0.5595\pm\mathrm{0.0081}$ & $0.0608\pm\mathrm{0.0190}$ & $26.14\pm\mathrm{0.71}$ & $0.5350\pm\mathrm{0.0147}$ \\
            PPA~\cite{struppek_plug_2022}     & $0.7760\pm\mathrm{0.0120}$ & $26.61\pm\mathrm{0.31}$ & $0.5080\pm\mathrm{0.0100}$ & $0.6708\pm\mathrm{0.0120}$ & $29.84\pm\mathrm{0.61}$ & $0.5260\pm\mathrm{0.0042}$ & $0.7480\pm\mathrm{0.0099}$ & $22.72\pm\mathrm{0.55}$ & $0.5448\pm\mathrm{0.0180}$ \\
            PLGMI~\cite{yuan_pseudo_2023}   & $0.3224\pm\mathrm{0.0224}$ & $50.26\pm\mathrm{2.24}$ & $0.4835\pm\mathrm{0.0056}$ & $0.4489\pm\mathrm{0.0251}$ & $49.62\pm\mathrm{0.33}$ & $0.4888\pm\mathrm{0.0106}$ & $0.5625\pm\mathrm{0.0055}$ & $57.52\pm\mathrm{1.31}$ & $0.4804\pm\mathrm{0.0116}$ \\
            IFGMI~\cite{qiu_closer_2024}   & $0.8241\pm\mathrm{0.0102}$ & $32.53\pm\mathrm{0.46}$ & $0.4873\pm\mathrm{0.0077}$ & $0.8062\pm\mathrm{0.0057}$ & $30.61\pm\mathrm{0.52}$ & $0.4753\pm\mathrm{0.0075}$ & $0.7157\pm\mathrm{0.0129}$ & $31.24\pm\mathrm{0.27}$ & $0.4507\pm\mathrm{0.0064}$ \\
            FGMIA~\cite{lu_fgmia_2025}   & $0.8749\pm\mathrm{0.0048}$ & $34.88\pm\mathrm{0.62}$ & $0.4306\pm\mathrm{0.0048}$ & $0.8719\pm\mathrm{0.0054}$ & $31.32\pm\mathrm{0.99}$ & $0.4631\pm\mathrm{0.0033}$ & $0.9116\pm\mathrm{0.0074}$ & $22.40\pm\mathrm{0.84}$ & $0.4046\pm\mathrm{0.0029}$ \\
            \midrule
            \textbf{SFMI (Ours)} & $\mathbf{0.8980}\pm\mathrm{0.0086}$ & $\mathbf{25.86}\pm\mathrm{0.21}$ & $\mathbf{0.4248}\pm\mathrm{0.0081}$ & $\mathbf{0.9103}\pm\mathrm{0.0073}$ & $\mathbf{27.61}\pm\mathrm{0.31}$ & $\mathbf{0.4419}\pm\mathrm{0.0077}$ & $\mathbf{0.9150}\pm\mathrm{0.0066}$ & $\mathbf{22.37}\pm\mathrm{0.74}$ & $\mathbf{0.3966}\pm\mathrm{0.0041}$ \\
            \bottomrule
        \end{tabular}}
        \vspace{2pt}
        \resizebox{\textwidth}{!}{\begin{tabular}{l|ccc|ccc|ccc}
            \toprule
            \multirow{2}{*}{\textbf{Method}} & \multicolumn{3}{c|}{\textbf{Target: ArcFace ($112\times112$)}} & \multicolumn{3}{c|}{\textbf{Target: MobileFaceNet ($112\times112$)}} & \multicolumn{3}{c}{\textbf{Target: ViT ($112\times112$)}} \\
            \cmidrule{2-10}
            & \textbf{ACC} $\uparrow$ & \textbf{FID} $\downarrow$ & \textbf{LPIPS} $\downarrow$ & \textbf{ACC} $\uparrow$ & \textbf{FID} $\downarrow$ & \textbf{LPIPS} $\downarrow$ & \textbf{ACC} $\uparrow$ & \textbf{FID} $\downarrow$ & \textbf{LPIPS} $\downarrow$ \\
            \midrule
            GMI~\cite{zhang_secret_2020}     & $0.0379\pm\mathrm{0.0258}$ & $29.89\pm\mathrm{0.58}$ & $0.5161\pm\mathrm{0.0093}$ & $0.1440\pm\mathrm{0.0158}$ & $26.50\pm\mathrm{0.14}$ & $0.5059\pm\mathrm{0.0010}$ & $0.0418\pm\mathrm{0.0065}$ & $29.71\pm\mathrm{1.09}$ & $0.5400\pm\mathrm{0.0068}$ \\
            PPA~\cite{struppek_plug_2022}     & $0.4191\pm\mathrm{0.0101}$ & $22.84\pm\mathrm{0.87}$ & $0.5594\pm\mathrm{0.0143}$ & $0.8036\pm\mathrm{0.0070}$ & $27.49\pm\mathrm{0.07}$ & $0.4549\pm\mathrm{0.0054}$ & $0.3698\pm\mathrm{0.0182}$ & $\mathbf{24.82}\pm\mathrm{0.38}$ & $0.5502\pm\mathrm{0.0082}$ \\
            PLGMI~\cite{yuan_pseudo_2023}   & $0.2357\pm\mathrm{0.0292}$ & $42.19\pm\mathrm{1.28}$ & $0.4587\pm\mathrm{0.0101}$ & $0.8648\pm\mathrm{0.0081}$ & $39.25\pm\mathrm{0.36}$ & $0.4383\pm\mathrm{0.0121}$ & $0.2555\pm\mathrm{0.0096}$ & $44.09\pm\mathrm{0.46}$ & $0.4511\pm\mathrm{0.0079}$ \\
            IFGMI~\cite{qiu_closer_2024}   & $0.7544\pm\mathrm{0.0263}$ & $35.79\pm\mathrm{0.51}$ & $0.4424\pm\mathrm{0.0017}$ & $0.8028\pm\mathrm{0.0073}$ & $34.84\pm\mathrm{0.37}$ & $0.4099\pm\mathrm{0.0045}$ & $0.7471\pm\mathrm{0.0115}$ & $33.57\pm\mathrm{0.49}$ & $0.4548\pm\mathrm{0.0047}$ \\
            FGMIA~\cite{lu_fgmia_2025}   & $0.8880\pm\mathrm{0.0087}$ & $\mathbf{21.72}\pm\mathrm{0.08}$ & $0.4088\pm\mathrm{0.0012}$ & $0.9009\pm\mathrm{0.0064}$ & $22.98\pm\mathrm{0.34}$ & $0.3884\pm\mathrm{0.0090}$ & $0.8074\pm\mathrm{0.0580}$ & $25.69\pm\mathrm{0.19}$ & $0.3952\pm\mathrm{0.0050}$ \\
            \midrule
            \textbf{SFMI (Ours)} & $\mathbf{0.9248}\pm\mathrm{0.0089}$ & $22.61\pm\mathrm{0.30}$ & $\mathbf{0.3874}\pm\mathrm{0.0059}$ & $\mathbf{0.9335}\pm\mathrm{0.0116}$ & $\mathbf{22.67}\pm\mathrm{0.74}$ & $\mathbf{0.3786}\pm\mathrm{0.0088}$ & $\mathbf{0.8735}\pm\mathrm{0.0147}$ & $25.61\pm\mathrm{0.65}$ & $\mathbf{0.3918}\pm\mathrm{0.0137}$ \\
            \bottomrule
        \end{tabular}}
    }
\end{table*}

Table~\ref{tab:quantitative_sota} reports the main quantitative comparison between SFMI and representative white-box MIA baselines on six target face recognition models, using ACC, FID, and LPIPS as complementary evaluation metrics. SFMI achieves the highest ACC on all evaluated target models and the lowest LPIPS across all targets, indicating strong identity recovery together with superior perceptual consistency. For FID, SFMI attains the best score on Face.evoLVe and remains highly competitive on the other target models, suggesting that the gain in attack success is not obtained at the cost of noticeable realism degradation. Taken together, these results demonstrate that SFMI offers a favorable trade-off between semantic fidelity and visual quality, and that this advantage generalizes across target models with different architectural and loss-design characteristics.

\begin{figure*}[t]
    \centering
    \includegraphics[width=0.98\linewidth]{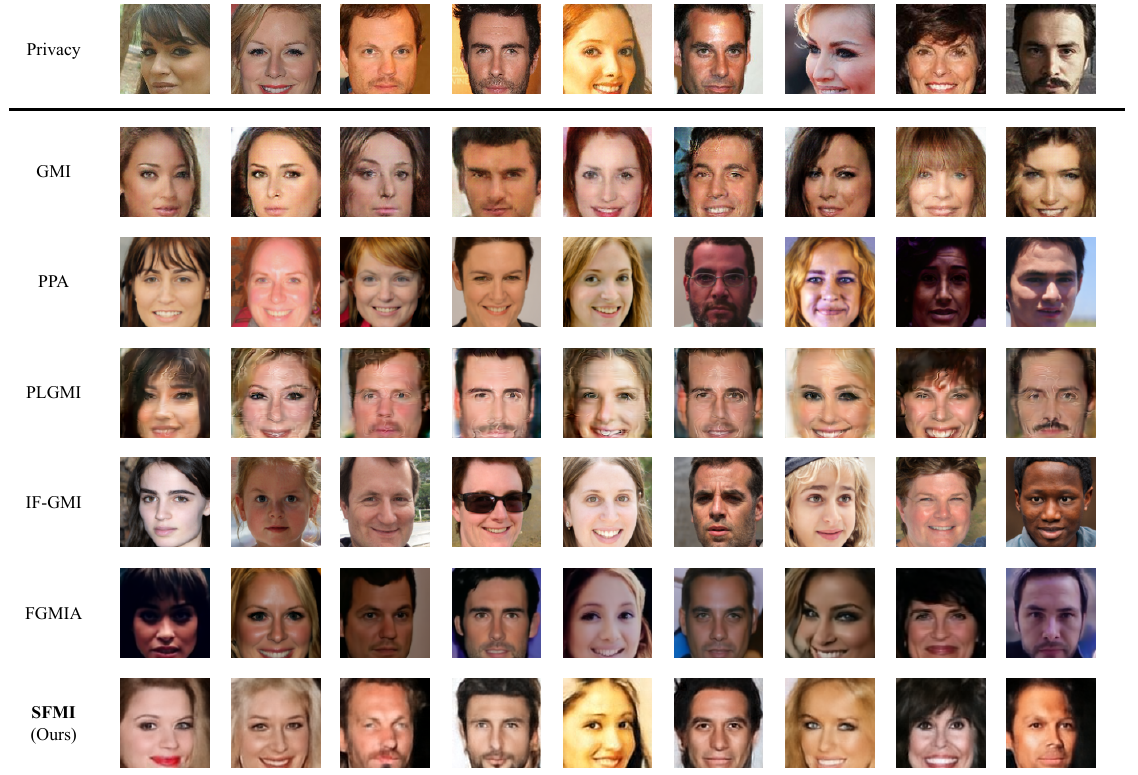}
    \caption{Qualitative visual comparison across MIA methods. The first row shows private ground-truth images, and rows 2--7 correspond to GMI, PPA, PLGMI, IFGMI, FGMIA, and SFMI, respectively.}
    \label{fig:qualitative_sota}
\end{figure*}

Figure~\ref{fig:qualitative_sota} further provides visual comparisons. The first row shows real private images, and rows 2--7 correspond to GMI, PPA, PLGMI, IFGMI, FGMIA, and SFMI, respectively. In most cases, SFMI preserves facial geometry and identity-related local traits more clearly, while reducing common artifacts such as blurry textures and unstable high-frequency noise observed in baseline reconstructions. We also observe that several previous methods are, to some extent, misled during optimization, which causes their reconstruction trajectories to drift away from target identities and leads to substantial inconsistencies with the corresponding private faces. These qualitative observations are well aligned with the quantitative trends in Table~\ref{tab:quantitative_sota}, further confirming the robustness and visual fidelity of SFMI across diverse identities.

\subsection{Ablation Studies}\label{subsec:ablation}
To rigorously validate the necessity and individual contributions of the core components within the proposed SFMI method, we conduct a comprehensive ablation study.
{We systematically isolate the effects of the generative prior model, the auxiliary dataset distribution, the Progressive Guidance Scheduler (PGS), the optimization loss design, and the input resolution.}
{Specifically, the variants include: replacing the FM prior with a DDPM prior paired with a DDIM sampler (\textit{DDPM prior + DDIM sampler}); training the prior on FFHQ-pub instead of CelebA-pub (\textit{w/ FFHQ Prior}); replacing PGS with a constant multiplier (\textit{w/ Constant Guidance}); replacing the MM loss with cross-entropy (\textit{w/ CE Loss (replace MM)}); and increasing the target input resolution (\textit{w/ 224 Resolution}).}
All ablation experiments are evaluated on the primary private dataset ($\mathcal{D}_{\text{priv}}$), with ArcFace as the main robust margin-based target and MobileFaceNet as an additional lightweight target for checking whether the component-level conclusions generalize across architectures. Following the standard MIA evaluation protocol, each target identity is attacked by sampling eight reconstructed images, and the reported metrics are averaged over all target identities rather than over a single example.
To further verify that the component-level conclusions are not specific to ArcFace, we additionally repeat the same key ablations on MobileFaceNet, as summarized in Table~\ref{tab:ablation_mobilefacenet}.

\begin{table}[h]
    \centering
    \caption{{Component and resolution ablations of SFMI on ArcFace.}}
    \label{tab:ablation_summary}
    \resizebox{\columnwidth}{!}{\begin{tabular}{l|ccc}
        \toprule
        \textbf{Variant} & \textbf{ACC} $\uparrow$ & \textbf{FID} $\downarrow$ & \textbf{LPIPS} $\downarrow$ \\
        \midrule
        DDPM prior + DDIM sampler & 0.1086 & 76.50 & 0.6084 \\
        w/ FFHQ Prior & 0.8995 & 24.67 & 0.4564 \\
        w/ Constant Guidance & 0.1658 & 80.66 & 0.5556 \\
        w/ CE Loss (replace MM) & 0.8842 & 23.99 & 0.4200 \\
        w/ 224 Resolution & 0.8599 & 25.21 & 0.4593 \\
        \midrule
        \textbf{SFMI (Full)} & \textbf{0.9248} & \textbf{22.61} & \textbf{0.3874} \\
        \bottomrule
    \end{tabular}}
\end{table}

\begin{table}[h]
    \centering
    \caption{{Component and resolution ablations of SFMI on MobileFaceNet.}}
    \label{tab:ablation_mobilefacenet}
    \resizebox{\columnwidth}{!}{\begin{tabular}{l|ccc}
        \toprule
        \textbf{Variant} & \textbf{ACC} $\uparrow$ & \textbf{FID} $\downarrow$ & \textbf{LPIPS} $\downarrow$ \\
        \midrule
        DDPM prior + DDIM sampler & 0.0879 & 62.70 & 0.5848 \\
        w/ FFHQ Prior & 0.8183 & 21.78 & 0.4390 \\
        w/ Constant Guidance & 0.0387 & 56.71 & 0.5958 \\
        w/ CE Loss (replace MM) & 0.8983 & 23.80 & 0.4291 \\
        w/ 224 Resolution & 0.8382 & 26.00 & 0.4237 \\
        \midrule
        \textbf{SFMI (Full)} & \textbf{0.9335} & \textbf{22.67} & \textbf{0.3786} \\
        \bottomrule
    \end{tabular}}
\end{table}

As observed in Table~\ref{tab:ablation_summary}, replacing the FM prior with a DDPM-trained prior and sampling it with DDIM (\textit{DDPM prior + DDIM sampler}) still results in a substantial degradation in semantic recovery. Following the original DDIM formulation, this baseline uses the DDPM noise-prediction model to estimate the denoised image $\hat{x}_0$ at each step, computes the target-model gradient through the predicted denoising path, and uses this gradient as guidance for DDIM sampling. {This variant provides a controlled proxy for diffusion-guided inversion under our common protocol; DiffMI additionally employs target-specific conditional fine-tuning~\cite{li2024model}.} Although DDIM removes step-wise sampling randomness, we observe that its guided trajectory is less robust and can more easily drift away from the natural image manifold. We attribute this to two factors: first, SFMI directly predicts the original image rather than relying on noise estimation, which is consistent with the denoising-oriented observation in~\cite{li_back_2025}; second, the DDPM prior induces a more irregular noise-to-face trajectory than Flow Matching, making target-class optimization more difficult. In contrast, the FM velocity field provides a smoother backbone for progressive identity-gradient steering.

Furthermore, to evaluate our method's sensitivity to distribution shifts, we train the prior on the identity-disjoint FFHQ-pub subset (\textit{w/ FFHQ Prior}) rather than the distributionally aligned CelebA-pub subset. Both public auxiliary subsets contain 30{,}000 images, so this comparison controls the auxiliary-data scale while changing only the public-prior distribution. The results indicate that while this dataset discrepancy introduces a minor degradation in performance, the overall attack remains highly effective. This demonstrates that SFMI does not strictly rely on matching the target's data distribution, exhibiting strong cross-distribution robustness.

Moreover, ablating the progressive schedule in favor of a static multiplier (\textit{w/ Constant Guidance}) severely compromises the practical usability of the method. Applying a rigid, constant gradient force uniformly across all integration steps irrecoverably corrupts the highly noisy initial states. This abrupt intervention forces the generative trajectory completely off the natural face manifold, leading to catastrophic optimization collapse. Consequently, the ACC plummets and the generated outputs suffer from severe, unrecognizable visual artifacts (evidenced by a spike in FID). This validates that the carefully modulated temporal dynamics of our progressive guidance are absolutely critical for maintaining the structural integrity of the generation while simultaneously injecting identity-specific features.

Regarding the loss design, replacing the margin-based MM loss with a standard cross-entropy objective (\textit{w/ CE Loss (replace MM)}) leads to a slight but consistent performance drop compared with the full setting. This observation highlights the superiority of MM for identity-level supervision and is broadly consistent with the findings reported in PLGMI~\cite{yuan_pseudo_2023}. Beyond validating our current design, it also suggests a promising direction for future work: developing more effective discriminative loss functions tailored to white-box model inversion.

To further examine higher-resolution behavior, we include a 224-resolution setting in the ablation tables. Scaling SFMI to a higher face-recognition input resolution requires a resolution-matched FM prior and the same detected/aligned/cropped FR preprocessing pipeline; thus, the main practical constraints are the target model's standardized input resolution and the computational cost of training and sampling the prior, rather than raw image size alone. Although the metrics moderately degrade at this higher resolution, the attack remains effective, indicating that SFMI is not restricted to low-resolution targets.

\subsection{Detailed Parameter Analysis of Progressive Guidance}\label{subsec:guidance_analysis}
Having established that a meticulously designed progressive guidance schedule is paramount to the method's overall usability, we now provide an in-depth parameter analysis of trajectory steering dynamics. We systematically evaluate seven variants of the guidance mechanism to dissect the functional necessity of both spatial direction alignment and the temporal warm-up, sustain, and annealing phases.

\begin{figure*}[t]
    \centering
    \includegraphics[width=0.98\linewidth]{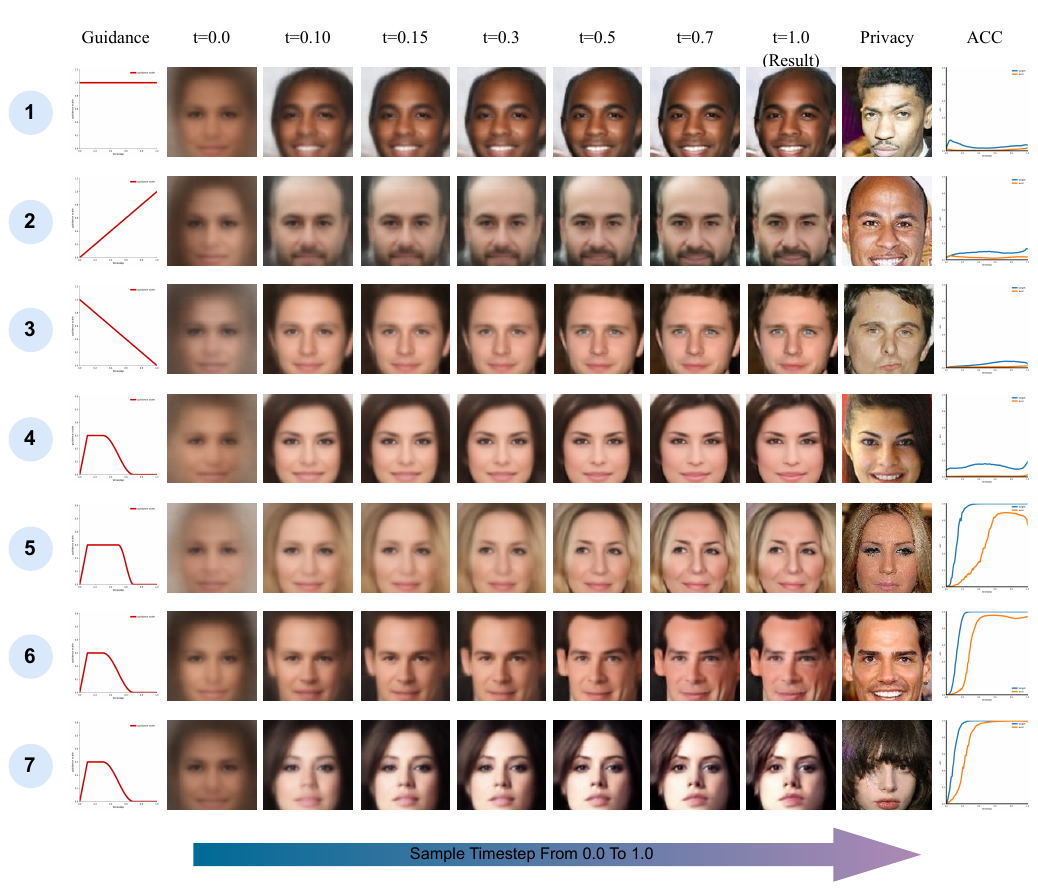}
    \caption{Detailed row-wise analysis of progressive guidance variants on the CosFace target model. Rows 1--7 correspond to constant guidance (1.0), linear increase (0.0$\rightarrow$1.0), linear decay (1.0$\rightarrow$0.0), unnormalized PGS, prolonged-peak PGS, standard PGS+CE loss, and standard PGS+MM loss (ours). For each row, the first column shows the guidance curve $\gamma(t)$, middle columns show intermediate trajectory states, and the final three columns report the reconstructed image, ground-truth face, and the ACC evolution curve over sampling steps. In the ACC curve, the blue line denotes ACC on the target model, and the yellow line denotes ACC on validation models.}
    \label{fig:guidance_variants}
\end{figure*}

To intuitively illustrate the impact of each hyperparameter and structural choice, Fig.~\ref{fig:guidance_variants} presents a comprehensive visual and quantitative matrix on the CosFace target model. The first column shows the mathematical curve of the guidance function $\gamma(t)$, and columns 2 through 8 visualize intermediate generation states from initial noise to a fully formed image. The final three columns report the reconstructed attack image, the corresponding real ground-truth face, and the ACC trajectory over guidance sampling steps, respectively.

A row-wise reading of Fig.~\ref{fig:guidance_variants} reveals several important behaviors. In Row 1, we remove PGS and apply a constant guidance coefficient of 1.0 without normalization; this seemingly simple guidance is too weak to effectively redirect the generation trajectory toward the target classifier. In Row 2, we again remove PGS and use a linear increase from 0.0 to 1.0 (also without normalization), but the guidance still fails to produce reliable identity steering. In Row 3, we use a linear decay from 1.0 to 0.0 without normalization and observe similarly weak control over facial trajectory evolution. In Row 4, we retain the PGS shape but remove normalization; although the schedule is more structured, the steering strength remains insufficient and reconstruction quality is still unsatisfactory. This weakness is also reflected by the last-column ACC curves, where both the blue target-model line and the yellow validation-model line remain relatively low or fluctuate unstably, indicating limited optimization effectiveness and poor transfer consistency.

Rows 5--7 further clarify how schedule design and loss design interact. In Row 5, based on standard PGS, we prolong the high-guidance stage and find that excessive late-stage forcing degrades performance on validation models, likely due to artifact accumulation in the refinement phase; correspondingly, the yellow curve drops in later steps and the blue-yellow gap widens, suggesting overfitting to short-term target gradients. In Row 6, standard PGS with CE loss already yields reasonably effective guidance and improved identity consistency, with both curves showing a clearer upward trend. In Row 7, standard PGS with our recommended MM loss achieves the best overall behavior: both target-model optimization and validation-model accuracy improve more stably, leading to the most favorable final reconstructions and the most consistent dual-curve progression.

The standard PGS configuration used in Eq.~\eqref{eq:guidance_schedule} is $M=0.3$, $t_0=0.1$, $t_1=0.3$, $t_2=0.7$, and $V_{\max}=1.0$. The scheduler progressively introduces identity guidance and then attenuates it before final refinement, thereby balancing effective steering toward the target identity with stable evolution along the learned face manifold. The PGS hyperparameters do not need to be reselected for every target model: by normalizing the identity gradient and scaling it relative to the current FM velocity, SFMI reduces sensitivity to variations in target-model gradient magnitude and yields relatively stable hyperparameter choices across architectures. In practice, we retain the standard relative phase allocation defined by $t_0$, $t_1$, and $t_2$, together with $V_{\max}$, and primarily adjust the global scale $M$, which is typically kept at or below 0.35 to preserve manifold stability.

\subsection{{Hyperparameter Sensitivity}}\label{subsec:hyperparameter_sensitivity}
\begin{figure}[t]
    \centering
    \includegraphics[width=\columnwidth]{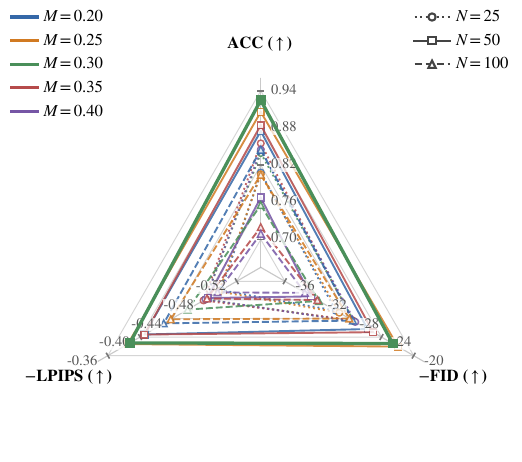}
    \caption{{Radar comparison of all 15 settings in Table~\ref{tab:hyperparameter_sensitivity}.} {Colors encode $M$ and line styles encode $N$; outward indicates better performance.}}
    \label{fig:hyperparameter_sensitivity}
\end{figure}

{Table~\ref{tab:hyperparameter_sensitivity} and Fig.~\ref{fig:hyperparameter_sensitivity} summarize the joint effects of guidance scale $M$ and sampling steps $N$ on ArcFace, with the phase thresholds and $V_{\max}$ fixed at the defaults in Sec.~\ref{subsec:exp_setup}.}
With the phase thresholds fixed, $M$ and $V_{\max}$ enter $\gamma(t)$ only through their product, so varying $M$ at $V_{\max}=1.0$ already sweeps the effective guidance amplitude.

\begin{table}[!ht]
    \centering
    \caption{{ArcFace sensitivity across 15 $M$--$N$ settings.} {Best values are bold.}}
    \label{tab:hyperparameter_sensitivity}
    \footnotesize
    \setlength{\tabcolsep}{6pt}
    \begin{tabular*}{\columnwidth}{@{\extracolsep{\fill}}cc|ccc@{}}
        \toprule
        $N$ & $M$ & ACC $\uparrow$ & FID $\downarrow$ & LPIPS $\downarrow$ \\
        \midrule
        \multirow{5}{*}{25}
        & 0.20 & 0.8080 & 29.68 & 0.5007 \\
        & 0.25 & 0.8052 & 29.24 & 0.4981 \\
        & 0.30 & 0.8395 & 27.43 & 0.4810 \\
        & 0.35 & 0.8551 & 27.63 & 0.4789 \\
        & 0.40 & 0.8449 & 27.26 & 0.4814 \\
        \midrule
        \multirow{5}{*}{50}
        & 0.20 & 0.8747 & 25.73 & 0.4056 \\
        & 0.25 & 0.9059 & \textbf{21.95} & \textbf{0.3849} \\
        & 0.30 & \textbf{0.9248} & 22.61 & 0.3874 \\
        & 0.35 & 0.8846 & 25.05 & 0.4054 \\
        & 0.40 & 0.7677 & 32.75 & 0.4840 \\
        \midrule
        \multirow{5}{*}{100}
        & 0.20 & 0.8449 & 27.55 & 0.4295 \\
        & 0.25 & 0.8045 & 28.00 & 0.4382 \\
        & 0.30 & 0.7556 & 31.77 & 0.4592 \\
        & 0.35 & 0.7200 & 31.96 & 0.4831 \\
        & 0.40 & 0.7093 & 33.52 & 0.4962 \\
        \bottomrule
    \end{tabular*}
\end{table}

{A small sampling budget tends to lower image quality, while excessive guidance can drive the trajectory away from the learned data manifold.}
When using $N\ne50$, we recommend reselecting the guidance scale $M$.
{We recommend $M=0.30$ and $N=50$ as the default configuration, keeping $M\leq0.35$ when adjusting the guidance scale at $N=50$.}

\subsection{{Defense Evaluation}}\label{subsec:defense_evaluation}
{We separately evaluate SFMI against two complementary defenses: output perturbation, a lightweight post-processing defense, and Bilateral Dependency Optimization (BiDO)~\cite{peng2022bilateral}, a training-phase model-inversion defense.}
{For output perturbation, we inject Gaussian noise with a standard deviation of 0.03 into the target-model output before computing attack guidance, without retraining the target model.}
{For BiDO, we train defended ArcFace and MobileFaceNet models on the same private-data split and attack them using the unchanged SFMI protocol.}

\begin{table}[t]
    \centering
    \caption{SFMI under output perturbation and BiDO on ArcFace and MobileFaceNet. OP denotes output perturbation.}
    \label{tab:defense_evaluation}
    \resizebox{\columnwidth}{!}{\begin{tabular}{llc|ccc}
        \toprule
        \textbf{Target} & \textbf{Defense setting} & \shortstack{\textbf{Target}\\\textbf{ACC} $\uparrow$} & \textbf{ACC} $\uparrow$ & \textbf{FID} $\downarrow$ & \textbf{LPIPS} $\downarrow$ \\
        \midrule
        \multirow{3}{*}{ArcFace}
        & None & 0.9796 & 0.9248 & 22.61 & 0.3874 \\
        & OP & 0.8557 & 0.8858 & 21.77 & 0.4320 \\
        & BiDO & 0.9110 & 0.8611 & 24.63 & 0.4291 \\
        \midrule
        \multirow{3}{*}{MobileFaceNet}
        & None & 0.9358 & 0.9335 & 22.67 & 0.3786 \\
        & OP & 0.8826 & 0.9012 & 23.01 & 0.4110 \\
        & BiDO & 0.8913 & 0.8572 & 26.39 & 0.3978 \\
        \bottomrule
    \end{tabular}}
\end{table}

{As shown in Table~\ref{tab:defense_evaluation}, BiDO reduces attack ACC more than output perturbation on both targets, at a measurable recognition-utility cost.}
{For ArcFace, BiDO reduces attack ACC from 0.9248 to 0.8611 (6.37 percentage points), while clean recognition accuracy decreases from 0.9796 to 0.9110 (6.86 percentage points).}
{For MobileFaceNet, attack ACC decreases from 0.9335 to 0.8572 (7.63 percentage points), while clean recognition accuracy decreases from 0.9358 to 0.8913 (4.45 percentage points).}

\section{Ethical Considerations}\label{sec:ethics}
This work studies model inversion attacks to audit privacy leakage in face recognition systems under a white-box, strong-access threat model. All experiments are conducted on publicly available research datasets, including CelebA and FFHQ, which are used only for academic evaluation in accordance with their dataset terms and without redistributing the original data. The qualitative reconstructions shown in the paper are included to demonstrate potential privacy leakage risks rather than to identify, profile, or target any individual. We treat publication as bounded academic disclosure of this privacy risk, acknowledge the dual-use nature of identity-reconstruction attacks, and frame SFMI as a privacy-auditing tool for researchers, developers, and system owners to motivate stronger privacy-preserving face recognition defenses, rather than as a practical guide for misuse.

\section{Conclusion}
\label{sec:conclusion}

In this paper, we introduced Steering Flow Model Inversion (SFMI), a method that reformulates white-box model inversion attacks as a deterministic trajectory-steering problem. By leveraging the continuous ODE formulation of Flow Matching, SFMI helps mitigate the optimization instability in highly non-convex GAN latent spaces and the stochastic disruption typical of standard diffusion models.
Furthermore, we designed a progressive gradient guidance mechanism that dynamically modulates adversarial force across integration steps, ensuring precise temporal and spatial alignment with the generative flow.
Extensive experiments demonstrate that SFMI delivers strong and competitive performance relative to state-of-the-art methods, particularly in attack accuracy and visual fidelity, while maintaining competitive distribution-level fidelity. SFMI also exhibits strong attack capability against robust margin-based face recognition architectures such as CosFace and ArcFace and maintains cross-distribution generalization.

SFMI still has limitations. {First, the sensitivity analysis and cross-target results support a common default configuration, while substantially different target models or data distributions may require further guidance adjustment.} Second, SFMI is a white-box method that requires full access to the target model including gradients, which corresponds to a relatively high-access setting in practical scenarios.
Third, SFMI depends on access to relevant public auxiliary face data, and its performance may degrade under larger domain shifts between the public and private data.
Fourth, SFMI is not intended to attack directly on IoT devices; instead, the deployed target model must first be extracted from the device and then attacked using external computing resources.
Fifth, cross-model ACC evaluation and LPIPS provide complementary evidence of identity recovery and perceptual similarity. Nevertheless, ACC remains a cross-model recognition proxy, and satisfying face recognition evaluators does not necessarily imply alignment with human identity perception.
Overall, SFMI provides a rigorous and effective tool for auditing privacy leakage in deep biometric systems. {In future work, we plan to investigate privacy-preserving face recognition models, weaker-access and black-box model inversion scenarios, adaptive guidance selection, stronger defensive paradigms against model inversion attacks, and model inversion methods that better align with human identity judgments.}

\bibliographystyle{IEEEtran}
\bibliography{ref}

\appendices
\section{Extended Qualitative Gallery of SFMI}
\label{app:qualitative_gallery}

\begin{figure*}[t]
    \centering
    \includegraphics[width=\textwidth]{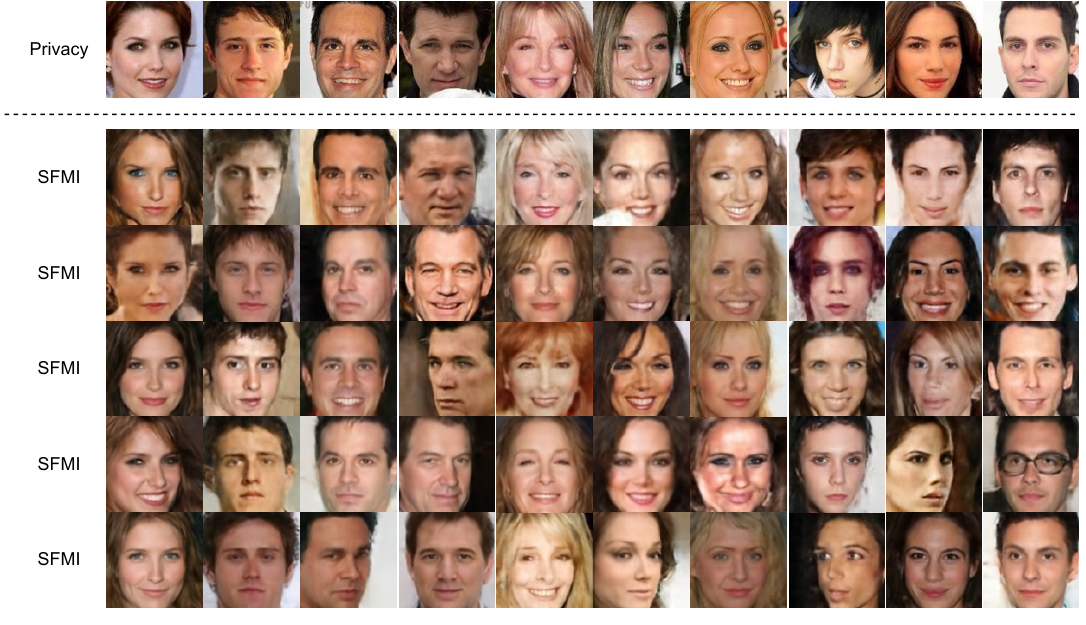}
    \caption{Extended SFMI reconstruction gallery on MobileFaceNet at $112\times112$ resolution. The first row shows private ground-truth images; rows 2--6 present the corresponding SFMI reconstructions obtained with the default attack hyperparameters.}
    \label{fig:appendix_qualitative_gallery}
\end{figure*}

To provide further intuition for the visual quality of SFMI, we present additional attack results in Fig.~\ref{fig:appendix_qualitative_gallery}. This appendix is presented as an extended SFMI reconstruction gallery under a fixed evaluation setting. The first row shows private ground-truth images, and rows 2--6 show the corresponding SFMI reconstructions. All attacks use the default hyperparameters described in Sec.~\ref{subsec:exp_setup}, with MobileFaceNet as the target model at $112\times112$ resolution.

\end{document}